\documentclass[journal]{IEEEtran}

\usepackage{amsmath,amssymb}
\usepackage{booktabs}
\usepackage{cite}
\usepackage{graphicx}
\usepackage{multirow}
\usepackage{array}
\usepackage{tabularx}
\usepackage{url}
\usepackage[hidelinks]{hyperref}
\usepackage{tikz}
\usepackage{pgfplots}
\pgfplotsset{compat=1.18}
\definecolor{echoblue}{HTML}{2455A6}
\definecolor{echopurple}{HTML}{7651B5}
\definecolor{echoteal}{HTML}{138F93}
\usetikzlibrary{arrows.meta,positioning,fit}

\title{ECHO: An Auditable Dual-Path Memory System for Long-Horizon Agents}

\author{Yu Qian\textsuperscript{1,*}, Hong Miao\textsuperscript{1,*},
Boyang Guo\textsuperscript{1,*}, Tingyi Jiang\textsuperscript{1},
Shan Zhao\textsuperscript{1}, Tianxing Le\textsuperscript{1},
Lintian Li\textsuperscript{1}, and Meng Liu\textsuperscript{1}%
\thanks{\textsuperscript{*}Equal contribution. \textsuperscript{1}XDream
Robotics, Shanghai, China. Correspondence to: Yu Qian
(\texttt{ethan@xdreamrobo.com}) and Meng Liu
(\texttt{meng.liu@xdreamrobo.com}).}}

\begin{document}
\maketitle

\begin{abstract}
Long-horizon agents require more than retrieval over stored conversations: they must preserve source evidence, revise outdated knowledge, and distinguish relevant memories from sufficient support for an answer. We present ECHO (Embodied Context \& History Orchestration), an auditable dual-path memory system. A fast path captures immutable events and serves multi-route recall, while asynchronous workers consolidate experience into revisable state and rebuild derived indexes. Source provenance, temporal authority, and answer realization are explicit boundaries. The service supports original-source packets with evidence-use checks and disables model-based reranking by default. For the latest reported accuracy, the authors confirm mixed retrieved inputs comprising original dialogue and memory summaries, answered by Qwen3-8B with NF4 quantization. In the latest author-supplied five-system aggregate evaluation, ECHO reports 56.27\% answer accuracy, 98.04\% source Hit@10, 93.58\% macro source Recall@5, and 13.92\% hallucination flags, with 19.54\,ms mean retrieval latency. Its reported accuracy exceeds Mem0 by 4.53 percentage points in this snapshot. The results are descriptive aggregates; matched question-level evaluation and generation latency remain to be established. We separate these observations from earlier published experiments and engineering regressions.
\end{abstract}

\section{Introduction}
An interactive agent accumulates information whose meaning and authority change over time. A user can revise a preference, correct an earlier statement, or refer to an event using a different name. A robot may observe an event before a later conversation explains it. Simply retrieving a semantically similar passage does not determine whether that passage is current, whether its speaker is the relevant person, or whether it contains every fact needed for the requested answer.

We study memory as a governed lifecycle from events to usable experience. ECHO separates the record of what happened from interpretations derived from it. The event record remains traceable; state can be revised; retrieval proposes evidence; temporal and provenance checks constrain its use. This decomposition permits an answer error to be localized to ingestion, discovery, state resolution, source packing, or generation rather than attributed to a single memory score.

The design is informed by the distinction between episodic and semantic memory \cite{tulving1972episodic} and complementary learning systems \cite{mcclelland1995cls,kumaran2016cls}. Rapid event capture and slower consolidation serve different computational roles. Temporal-context models motivate contextual retrieval \cite{howard2002temporal}, while cognitive-control accounts motivate task-dependent evidence selection \cite{miller2001pfc}. These are functional design analogies, not claims of biological equivalence or evidence of human-like intelligence.

Existing approaches contribute memory streams and reflection, virtual context management, linked notes, extraction and updates, and temporal graphs \cite{park2023generative,packer2023memgpt,xu2025amem,chhikara2025mem0,rasmussen2025zep}. ECHO does not claim that any one of these mechanisms is new. Its focus is their integration under explicit authority boundaries: derived indexes locate candidates, the state ledger resolves revision eligibility, source records support interpretation, and a separately evaluated reader produces an answer.

This revision describes the current dual-path service and incorporates the latest five-system aggregate evaluation. Its provenance and relationship to earlier published experiments are made explicit in Sections~\ref{sec:eval} and~\ref{sec:results}.

Our contributions are threefold. First, we specify a dual-path architecture that decouples durable experience capture from asynchronous consolidation while retaining source and revision identity. Second, we make original-source evidence packing, evidence-use checks, and failure boundaries explicit in the answering pipeline. Third, we report the latest supplied aggregate comparison with ECHO, Mem0, A-MEM, Letta, and LangMem, and distinguish descriptive rankings from the matched, question-level evidence needed for confirmatory claims.
\section{Problem Formulation}

\label{sec:problem}

\subsection{Events, States, and Two Time Axes}

Let an interaction history be an append-only event stream
$E=(e_1,\ldots,e_T)$. Each event contains immutable source identity, speaker or
actor, observed text, source order, and available timestamps. Projection maps
events to typed propositions and state revisions. A state record is

\begin{equation}
s=(k,v,\tau_v,\tau_t,\rho,\pi,g,\sigma),
\label{eq:state}
\end{equation}

where $k$ is a typed entity--relation key, $v$ a value, $\tau_v$ its valid-time
interval, $\tau_t$ the transaction or known-time interval, $\rho$ the revision
relation, $\pi$ immutable provenance, $g$ a ledger generation, and $\sigma$ a
status such as active, superseded, revoked, or unresolved. Valid time denotes
when a proposition is true in the modeled world; transaction time denotes when
the system records it. This separation follows established bitemporal database
semantics \cite{torp2000timestamping}.

For a query view $t=(t_v,t_t)$, ledger resolution is

\begin{equation}
\mathcal{S}(k,t)=\{s: s.k=k,\;t_v\in s.\tau_v,\;t_t\in s.\tau_t,
\;s.g\leq g_{\mathrm{visible}}\}.
\label{eq:resolution}
\end{equation}

If $\mathcal{S}$ contains one supported active revision, that state can be
answered. If it is empty, evidence is missing. If it contains incompatible
unresolved revisions, the system must not pick the most similar one; it either
discloses the minimal conflict set when requested or abstains.

We make this authority boundary explicit with a three-valued resolver:

\begin{equation}
\operatorname{Resolve}(k,t)=
\begin{cases}
s, & \mathcal{S}_{\mathrm{adm}}(k,t)=\{s\},\\
\bot_{\mathrm{miss}}, & \mathcal{S}_{\mathrm{adm}}(k,t)=\varnothing,\\
\bot_{\mathrm{conflict}}, & |\mathcal{S}_{\mathrm{adm}}(k,t)|>1,
\end{cases}
\label{eq:resolve-decision}
\end{equation}

where admissibility requires compatible valid time, known time, revision status,
provenance, and visible ledger generation. Semantic similarity is deliberately
absent from Eq.~\eqref{eq:resolve-decision}: it may discover a state, but it
cannot make that state authoritative.

\subsection{Evidence and Answer Operations}

A retrieval hit is not necessarily sufficient evidence. Let $u$ be an evidence
unit and $\Pi(u)$ its source lineage. For a selected state or derived result,
the evidence closure $\Gamma(u)$ contains the originating events, required
revision links, and dependencies needed to verify the result. Candidate
discovery first unions route-specific results,

\begin{align}
\mathcal{H}(q) &= \bigcup_{r\in\mathcal{R}}\operatorname{Top}_{M_r}
  \bigl(\mathcal{H}_r(q)\bigr),\\
F(u\mid q) &= \sum_{r\in\mathcal{R}}\alpha_r\phi_r(u,q)
  +\lambda\,\operatorname{Support}(\operatorname{session}(u)),
\label{eq:candidate-fusion}
\end{align}

where routes include lexical, semantic, typed-state, and neighborhood
discovery. The fusion score $F$ orders inspection only; currentness remains the
output of Eq.~\eqref{eq:resolve-decision}. Evidence packing is an
operation-conditioned atomic-closure problem,

\begin{align}
\mathbf{x}^{\star}
&=\arg\max_{\mathbf{x}\in\{0,1\}^{|\mathcal{H}|}}
  \sum_{u\in\mathcal{H}}x_u\,U(u\mid q,o),\\
\text{s.t.}\quad
&\sum_{u\in\mathcal{H}}x_u\,\operatorname{tokens}(\Gamma(u))\leq B,
\qquad C=\bigcup_{u:x_u^{\star}=1}\Gamma(u),
\label{eq:closure}
\end{align}

for utility-conditioned evidence units $\mathcal{H}$ and budget $B$. A packer
may defer an atomic closure that does not fit, but it cannot silently include
only the conclusion while dropping its required support.

The public question induces an operation such as \textsc{Fact}, \textsc{Count},
\textsc{List}, \textsc{TemporalLookup}, or \textsc{TemporalArithmetic}. The
operation determines which evidence must be complete and which answer surface
is required. Importantly, internal completeness is not equivalent to visible
completeness.

\subsection{Objective and Claim Taxonomy}

For question $q$, frozen context $C$, internal derivation $z$, and surface
contract $h$, the reader computes

\begin{align}
z &= D(q,C,o), \\
a &= R(q,z,h).
\label{eq:derive_render}
\end{align}

$D$ may enumerate members or retain temporal endpoints. $R$ emits only the
answer requested by $q$. We optimize all-row strict answer correctness while
also reporting support recall, abstention behavior, token use, and latency.
Retrieval metrics diagnose $C$; they never substitute for the correctness of
$a$.

The resulting system objective is constrained rather than a single retrieval
score. For rows $i=1,\ldots,n$,

\begin{equation}
\begin{split}
\mathcal{J}={}&\frac{1}{n}\sum_{i=1}^{n}\mathbf{1}[a_i\equiv y_i]
-\lambda_s L_{\mathrm{stale}}-\lambda_c L_{\mathrm{conflict}}\\
&-\lambda_f L_{\mathrm{false\_abstain}}-\lambda_b\,\overline{B},
\end{split}
\label{eq:system-objective}
\end{equation}

subject to provenance closure and generation-fencing invariants. The auxiliary
terms expose safety and cost failures that a high Hit@$k$ can hide.

These equations define the architectural contract. They do not imply that every retrieval benchmark exercises every resolver, closure, or action policy. In particular, Eq.~\eqref{eq:closure} is a budgeted selection specification, not a claim that the deployed packer solves an exact combinatorial optimum. We distinguish implementation contracts, engineering tests, aggregate evaluation snapshots, and matched confirmatory experiments throughout the paper.

\section{Dual-Path Memory Architecture}
\label{sec:architecture}
\subsection{A Stable Memory Core and Pluggable Interpretation}
The stable core owns event identity, subject scope, source lineage, revisions, and lifecycle state. Model-dependent components interpret events, propose consolidation, and assist recall or reading through explicit interfaces. This separation allows a model or retrieval backend to change without silently redefining the source of truth. Figure~\ref{fig:architecture} shows the current architectural decomposition and the intended feedback path.

\begin{figure*}[t]
\centering
\begin{tikzpicture}[x=1cm,y=1cm,font=\sffamily\scriptsize,
box/.style={draw=echopurple!65,fill=echopurple!5,rounded corners=3pt,align=center,text width=2.6cm,minimum height=1.02cm,inner sep=4pt},
slow/.style={box,draw=echoteal!70,fill=echoteal!5},
flow/.style={-{Latex[length=2mm]},thick,draw=echopurple},
async/.style={-{Latex[length=2mm]},thick,dashed,draw=echoteal}]
\node[anchor=west,font=\sffamily\bfseries\small,text=echopurple] at (-1.48,1.05) {FAST PATH \quad Capture, retrieve, and respond};
\node[box] (input) at (0,0) {\textbf{1. Input}\\Dialogue / perception\\API events / time};
\node[box] (append) at (3.25,0) {\textbf{2. Encode \& append}\\Immutable source record\\Actor / order / timestamp};
\node[box] (recall) at (6.5,0) {\textbf{3. Multi-route recall}\\Lexical / dense / entity\\Temporal and scope filters};
\node[box] (reason) at (9.75,0) {\textbf{4. Resolve \& read}\\Revision and source checks\\Original-source packets};
\node[box] (act) at (13,0) {\textbf{5. Output}\\Answer / abstention\\Action and feedback hooks};
\draw[flow] (input)--(append);\draw[flow] (append)--(recall);\draw[flow] (recall)--(reason);\draw[flow] (reason)--(act);
\node[box,text width=5.7cm,minimum height=.62cm] (buffer) at (3.25,-1.5) {\textbf{Durable evidence + projection job}\\Append-only source identity, content, time, scope, and hash};
\draw[flow] (append)--(buffer);
\node[anchor=west,font=\sffamily\bfseries\small,text=echoteal] at (-1.48,-2.8) {SLOW PATH \quad Consolidate, revise, and maintain};
\node[slow] (consolidate) at (0,-3.8) {\textbf{1. Consolidate}\\Session / event grouping\\Typed interpretation};
\node[slow] (revise) at (3.25,-3.8) {\textbf{2. Revise}\\ADD / REVISE / NOOP\\Source-linked changes};
\node[slow] (ledger) at (6.5,-3.8) {\textbf{3. Record}\\Persistent memory ledger\\Valid / known-time views};
\node[slow] (index) at (9.75,-3.8) {\textbf{4. Rebuild indexes}\\Lexical / vector / graph\\Derived and replaceable};
\node[slow] (govern) at (13,-3.8) {\textbf{5. Audit \& govern}\\Retention / quality gates\\Policy-update boundary};
\draw[async] (buffer.west)--(-1.7,-1.5)--(-1.7,-3.8)--(consolidate.west);
\draw[async] (consolidate)--(revise);\draw[async] (revise)--(ledger);\draw[async] (ledger)--(index);\draw[async] (index)--(govern);
\draw[async] (index.north)--(9.75,-2.2)--(6.5,-2.2)--(recall.south);
\node[align=center,text=black!65,font=\sffamily\tiny] at (10.9,-1.28) {Solid: request path\qquad Dashed: asynchronous maintenance\\Evolution and action benefits require separate evaluation};
\end{tikzpicture}
\caption{ECHO dual-path architecture, redrawn from the authors' system diagram. Durable capture and asynchronous projection are separate completion events. Derived indexes support discovery; revision and provenance checks constrain evidence use. The figure describes system responsibilities, not a claim that autonomous policy evolution or embodied action has been validated by the QA results.}
\label{fig:architecture}
\end{figure*}

\subsection{Fast Path: Durable Capture and Bounded Recall}
Inputs can carry dialogue, perception-derived records, API events, and time or context metadata. A source record preserves its actor or speaker, source order, timestamp, and scope. The immediate buffer supports bounded current context; durable ingestion records both evidence and a projection job. A successful write gate means that this evidence/job pair has been persisted. It does not promise that a new semantic memory has already been formed.

Scope includes subject, robot, and session identity. These fields constrain admissible memory before relevance ranking. Entity resolution must also preserve who made a statement and whom it concerns: a preference expressed by one speaker must not be silently assigned to another. Deterministic identifiers make repeated identical writes idempotent, while reuse of an identifier with different content is rejected.

Recall combines lexical, dense, entity, recent-context, and temporal signals. Lexical matching preserves exact anchors, while dense retrieval supports paraphrases \cite{robertson2009bm25}. Route fusion ranks candidates for inspection. Neither a high vector score nor an index timestamp makes a candidate authoritative; eligibility depends on the appropriate revision and time view. A recall identifier and route diagnostics support subsequent examination of missing evidence or degraded routes.

\subsection{Slow Path: Projection and Revision}
Asynchronous workers interpret persisted evidence through a projection run. The result may add a memory, revise an existing one, or make no change. A NOOP can be correct when an event contains no durable information or duplicates existing knowledge. Source evidence is retained independently of the projection outcome, so a failed or revised interpretation need not destroy the event history.

Projected state and relation views reference their source records, following database provenance principles \cite{cheney2009provenance}. Revisions preserve earlier interpretations instead of overwriting their lineage. Valid time captures when a proposition applies in the modeled world; known time captures when it entered the system. These are distinct from the order in which a worker finishes a job. Historical queries must use the requested view rather than a recent-context shortcut that introduces later information.

Index maintenance follows the ledger. Lexical, vector, and graph structures can be rebuilt from persisted state and provenance. This makes repair an explicit operation rather than a hidden change to the meaning of an existing memory. Retention, conflict detection, and quality monitoring belong to the governance boundary; learned policy changes require additional validation before promotion.

\subsection{Original-Source Packing and Evidence-Use Checks}
The current default disables model-based reranking. Retrieved memory links are resolved to original-source records for the reader. The packer preserves content and source identity, retains actor and temporal metadata, deduplicates repeated records, and records exclusions caused by the context budget. It must not replace omitted support with a fabricated summary or use a reference answer to fill a gap.

The packet separates the public question, retrieved evidence, and output contract. Evidence-use checks verify that cited identifiers refer to visible records and that required structured fields are present. Such checks are useful engineering constraints, but a valid whole-record citation is not a proof that every assertion is entailed by that record. A separate support judgment is therefore needed. Source-level retrieval metrics likewise do not prove that the exact answer-bearing fact reached the reader.

For operations such as counting or temporal arithmetic, internal evidence requirements can be richer than the requested answer surface. A count requires a complete, deduplicated set; a duration requires two supported endpoints. The final response should express the requested quantity rather than expose unrelated intermediate material. If required support is absent or contradictory revisions remain unresolved, the reader should abstain. This is an evidence-dependent rejection policy, related to selective prediction \cite{geifman2017selective}, rather than an inference from semantic similarity alone.

\subsection{Controlled Evolution as a Research Boundary}
A slow path creates a place for feedback-driven improvement, but feedback collection is not itself evidence of autonomous learning. A candidate policy should be versioned, evaluated on histories not used to construct it, constrained by correctness and resource budgets, and reversible. Formally, if $p$ is a proposed policy and $\pi$ the deployed policy, promotion requires
\begin{equation}
\operatorname{Promote}(p)=\mathbf{1}[\Delta Q\geq\delta]\,
\mathbf{1}[\Delta H\leq0]\,\mathbf{1}[C(p)\leq C_{\max}],
\end{equation}
where $Q$ denotes a prespecified quality metric, $H$ unsupported-answer risk, and $C$ cost. Thresholds and evaluation histories must be fixed before observing outcomes. This is a proposed validation contract; the aggregate experiment below contains no policy-promotion or embodied-action trial.

\section{Implementation and Engineering Validation}
The current service is implemented in C++20 in a ROS 2 Jazzy-oriented codebase. DuckDB is the authoritative store. The graph backend and vector indexes are derived structures, not independent authorities for changing a fact. Model-facing interpretation is pluggable, while source identity, lifecycle records, and revision semantics belong to the stable service core. This architectural account follows the repository snapshot inspected for this revision; it is not a substitute for the exact revision manifest of each benchmark run.

Engineering validation addresses native ingestion and recall, source traceability, speaker/entity/time fields, UTF-8 handling, JSON schemas, context budgets, failure blocking, and GPU resource protection. Each check has a different scope. Schema tests establish whether a response can be consumed; evidence tests establish whether a source pointer resolves; budget tests establish whether a packet fits; resource tests establish whether a job is admitted safely. None establishes that an answer is correct.

\begin{table}[t]
\caption{Engineering contracts and their evidential limits.}
\label{tab:engineering}
\centering\scriptsize
\begin{tabularx}{\columnwidth}{p{.25\columnwidth}XX}
\toprule
Boundary & Required check & Does not establish \\
\midrule
Ingestion & Durable source/job identity; idempotence & Correct semantic projection \\
Projection & Schema, provenance, revision links & True or useful interpretation \\
Recall & Scope, time view, source trace & Complete answer evidence \\
Packing & UTF-8, JSON, budget, omission trace & Entailment of each claim \\
Generation & Parseable response, visible citation IDs & Answer correctness \\
Resources & Admission, failure blocking, release & Production latency or quality \\
\bottomrule
\end{tabularx}
\end{table}

CPU mocks and synthetic regressions exercise these contracts without a real reader. They must be reported separately from model-based QA. A failed schema, broken provenance link, or incomplete judge batch should stop the affected stage and preserve its diagnostic record. Retries and repaired attempts must be distinguished from first-attempt quality and latency rather than silently replacing failures.

\section{Evaluation Design and Evidence Scope}
\label{sec:eval}
\subsection{Research Questions and Intended Benchmark Suite}
We ask whether the current system retrieves useful sources, converts them into correct supported answers, and does so with tractable latency. These questions require separate measures. LoCoMo tests long conversational histories \cite{maharana2024locomo}; LongMemEval tests extraction, updates, temporal and multi-session reasoning \cite{wu2025longmemeval}; BEAM extends long-horizon memory assessment to larger histories \cite{tavakoli2025beam}. The intended comparison cohort is ECHO, Mem0, A-MEM, Letta, and LangMem.

The latest numerical input to this revision is an author-supplied spreadsheet containing one aggregate row per system. In a subsequent protocol clarification on 19 October 2026, the authors confirmed mixed original-dialogue and memory-summary inputs and a Qwen3-8B reader with NF4 quantization for ECHO's reported 56.27\% accuracy. The authors identify it as the latest run and superseding the earlier local continuation report. It has no per-dataset sheet, question list, denominator, or model/runtime manifest. Although the surrounding evaluation work includes a 201-question development set, we cannot infer the latest table's weighting or binary success counts from its percentages. We therefore preserve its reported proportions, do not assign them a reconstructed denominator, and do not label them as separate LoCoMo, LongMemEval, or BEAM scores.

\subsection{Evaluation Pipeline}
The evaluation pipeline separates native memory writing, multi-route retrieval, evidence packing, reader generation, and independent grading through an official subscription-based model endpoint. For the latest ECHO accuracy, the packed reader evidence combines original dialogue and memory summaries; the answerer is Qwen3-8B with NF4 quantization. This mixed-input setting must be distinguished from the original-source-only reader path described in the architecture. Correctness and support are different judgments: the former compares a candidate against the requested answer, while the latter asks whether personalized factual claims are supported by the actual visible evidence. An abstention may be non-hallucinatory yet incorrect on an answerable question.

The independent grading stage is separate from the answerer, not an official benchmark certification. The confirmed reader model is not the independent judge or the embedding model. The latest judge model identifier, rubric, decoding configuration, and provider version still require a run-specific manifest; older run metadata cannot establish these fields for the latest aggregate.

\subsection{Reader Inputs and Model for the Latest Accuracy}
\label{sec:reader-repro}
The reported ECHO accuracy of 56.27\% is a downstream QA result under a particular retrieval-and-reader configuration. It is not an intrinsic accuracy of the memory store or a retrieval-only measure. The authors confirmed two reader-visible evidence types: \emph{original dialogue} recovered through retrieval and \emph{memory summaries} produced from the stored history. Qwen3-8B, deployed with NF4 quantization, consumes their mixture together with the question and generates the candidate answer. These statements are author-confirmed protocol metadata for the latest spreadsheet, rather than an inference from an earlier run.

The raw-dialogue component supplies the original conversational evidence; the summary component supplies the retrieved, condensed memory representation. They should remain distinguishable in a released prompt and trace, because a summary can omit qualifiers or contain a projection error. The lexical, dense, entity, and temporal routes described in Section~\ref{sec:architecture} explain candidate discovery at the architecture level. The latest aggregate does not identify the enabled routes or the exact mixture policy for this run. In particular, Hit@10 and Recall@5 are evaluation cutoffs: they do not establish that the reader received exactly ten or five dialogue records, or a particular number of summaries.

\begin{table}[t]
\caption{Reader configuration associated with the latest ECHO accuracy. Confirmed entries were supplied by the authors on 19 October 2026; unspecified entries are not filled using older experiments.}
\label{tab:reader-config}
\centering\footnotesize
\begin{tabularx}{\columnwidth}{p{.28\columnwidth}X}
\toprule
Field & Latest ECHO aggregate \\
\midrule
Answer accuracy & 56.27\% \\
Reader evidence & Mixed retrieved original dialogue and memory summaries \\
Answer model & Qwen3-8B; NF4 quantization \\
Evidence selection & Per-type top-$k$, order, deduplication and truncation policy not supplied \\
Reader runtime & Weight/tokenizer revision, chat template, token limits, thinking mode, sampling settings and seed not supplied \\
Independent judge & Separate grading stage; exact model/version and rubric not supplied for this snapshot \\
\bottomrule
\end{tabularx}
\end{table}

For reproduction, the release must bind every question to the \emph{exact rendered mixed-input prompt}: ordered raw-dialogue and summary blocks, source IDs and summary-to-source links, retained speaker/time fields, omitted records, and token counts. The prompt template, summary-generation configuration, model and tokenizer revisions, NF4 loading settings, inference backend, and decoding parameters must accompany the output and grading trace. Reference answers belong only to correctness grading and must not enter retrieval, packing, or reader generation. Support grading should inspect what the reader actually saw, without filling omitted evidence from the full history.

The confirmation above identifies the reader and evidence types, but does not establish a common reader, quantization, or context budget for all five systems. Consequently, the table compares reported system pipelines; it cannot isolate memory quality under a shared reader. We do not transfer the temperatures, token limits, source-only packing rules, or judge identifiers of the separate 201-question continuation run to the latest 56.27\% result. A machine-readable disclosure accompanying this revision records confirmed fields and leaves unavailable run settings explicitly unspecified.

\subsection{Metric Definitions}
Let $G_i$ denote the set of relevant source units and $R_{i,k}$ the source units returned within the first $k$ ranks. With $g_i=|G_i|>0$, source coverage is
\begin{align}
\operatorname{Hit@}k &= n^{-1}\sum_i\mathbf{1}[|R_{i,k}\cap G_i|>0],\\
\operatorname{Recall}_{\rm macro}@k &= n^{-1}\sum_i\frac{|R_{i,k}\cap G_i|}{g_i},\\
\operatorname{Recall}_{\rm micro}@k &= \frac{\sum_i|R_{i,k}\cap G_i|}{\sum_i g_i}.
\end{align}
Source P@10 measures relevant-source precision at cutoff 10; MRR@10 averages reciprocal first-relevant rank, with zero for no relevant result in the top ten. A reproducible run must define source granularity, deduplication, treatment of fewer than ten results, and handling of questions without annotated evidence. The spreadsheet identifies the source metrics and cutoffs but does not supply all of these implementation details.

Answer accuracy and hallucination flags are reported independently. Their denominators and aggregation weights must be stated before pooling datasets. Retrieval latency is the time for the recorded retrieval stage; generation latency and wall-clock question-to-answer latency are additional quantities. Mean and percentile retrieval statistics cannot be presented as generation or end-to-end latency.

\subsection{Comparability and Statistical Requirements}
A matched comparison should freeze question IDs, corpus revisions, system versions, ingestion policies, reader and judge configurations, context budgets, retrieval cutoffs, and failure handling. A baseline without observable source lineage should receive ``not observable'' for provenance metrics rather than a synthetic zero. The existence of source scores in an aggregate table does not alone verify that different systems used equivalent provenance mappings.

For a future question-level rerun, paired accuracy differences should be accompanied by history-cluster bootstrap intervals; exact McNemar tests can describe paired discordances with clustering limitations made explicit. Such tests cannot be recovered from five aggregate rows. We report no invented error bars, confidence intervals, significance claims, or ablation effects for the latest snapshot. A separate, frozen confirmatory split remains necessary for a general superiority claim.

\section{Latest Aggregate Results}
\label{sec:results}
\begin{table*}[t]
\caption{Latest aggregate evaluation. Percentages except MRR; higher is better except hallucination flags. Source metrics retain the supplied labels and cutoffs. Protocol scope is specified in Section~\ref{sec:eval}.}
\label{tab:quality}
\centering\small\setlength{\tabcolsep}{9pt}
\begin{tabular}{lrrrrrrr}
\toprule
System & Accuracy & Hallucination & Source P@10 & Hit@10 & Recall@5 macro & Recall@5 micro & MRR@10 \\
\midrule
ECHO & \textbf{56.27} & \textbf{13.92} & \textbf{54.98} & \textbf{98.04} & \textbf{93.58} & \textbf{86.69} & \textbf{0.8943} \\
Mem0 & 51.74 & 16.92 & 48.21 & 96.02 & 87.81 & 84.86 & 0.8320 \\
A-MEM & 42.79 & 26.37 & 31.09 & 96.02 & 85.66 & 84.51 & 0.8113 \\
Letta & 39.30 & 28.36 & 29.80 & 94.03 & 83.25 & 81.69 & 0.7986 \\
LangMem & 20.90 & 23.88 & 19.00 & 71.64 & 53.81 & 52.82 & 0.4814 \\
\bottomrule
\end{tabular}
\end{table*}

\subsection{Answer Quality and Support}
With the author-confirmed mixed original-dialogue/summary inputs and Qwen3-8B NF4 reader (Section~\ref{sec:reader-repro}), ECHO has the highest reported answer accuracy in Table~\ref{tab:quality}: 56.27\%, compared with 51.74\% for Mem0, 42.79\% for A-MEM, 39.30\% for Letta, and 20.90\% for LangMem. The ECHO--Mem0 difference is 4.53 percentage points. This is a within-table descriptive ranking; statistical significance and causal attribution cannot be established from the available aggregate record.

The ECHO hallucination-flag rate is 13.92\%, versus 16.92\% for Mem0, a reduction of 3.00 percentage points. A-MEM, Letta, and LangMem report 26.37\%, 28.36\%, and 23.88\%, respectively. Lower hallucination does not by itself imply better answer completion: an unsupported abstention can reduce unsupported assertions while still being wrong. Accuracy and hallucination therefore remain separate columns rather than being collapsed into a single quality score.

\subsection{Source Retrieval and the Answer Gap}
ECHO reports 98.04\% source Hit@10, 54.98\% source P@10, and 0.8943 MRR@10. Macro source Recall@5 is 93.58\%, while micro recall is 86.69\%. The 6.89-point macro--micro difference is consistent with harder coverage for questions requiring more source units, but the aggregate table cannot identify which question types cause it. That hypothesis requires per-question source counts and error analysis.

High source retrieval and moderate answer accuracy describe different stages. A relevant source can contain distracting material, lack the necessary sentence in a visible packet, concern the wrong entity, or require a temporal or multi-source inference that the reader fails to perform. Without paired retrieval/answer traces, the difference between Hit@10 and answer accuracy cannot be interpreted as a measured conditional reader failure rate. Figure~\ref{fig:comparison} keeps the quality, coverage, and latency axes separate.

\begin{figure*}[t]
\centering
\begin{tikzpicture}
\begin{axis}[width=.325\textwidth,height=4.65cm,ybar,bar shift=0pt,bar width=14pt,ymin=0,ymax=65,
xtick={1,2,3,4,5},xticklabels={ECHO,Mem0,A-MEM,Letta,LangMem},
xticklabel style={font=\tiny,rotate=25,anchor=east},yticklabel style={font=\scriptsize},
ylabel={Percent},ylabel style={font=\scriptsize},title={(a) Answer accuracy},title style={font=\small},
ymajorgrids=true,grid style={black!10},axis line style={black!30},
nodes near coords,point meta=y,nodes near coords style={font=\tiny,/pgf/number format/fixed,/pgf/number format/precision=1},
enlarge x limits=.16]
\addplot[fill=echoblue!35,draw=none] coordinates {(1,56.268657) (2,51.741294) (3,42.786070) (4,39.303483) (5,20.895522)};
\addplot[fill=echoblue,draw=none,nodes near coords={}] coordinates {(1,56.268657)};
\end{axis}\end{tikzpicture}
\begin{tikzpicture}
\begin{axis}[width=.325\textwidth,height=4.65cm,ybar,bar shift=0pt,bar width=14pt,ymin=0,ymax=110,
xtick={1,2,3,4,5},xticklabels={ECHO,Mem0,A-MEM,Letta,LangMem},
xticklabel style={font=\tiny,rotate=25,anchor=east},yticklabel style={font=\scriptsize},
ylabel={Percent},ylabel style={font=\scriptsize},title={(b) Source Hit@10},title style={font=\small},
ymajorgrids=true,grid style={black!10},axis line style={black!30},
nodes near coords,point meta=y,nodes near coords style={font=\tiny,/pgf/number format/fixed,/pgf/number format/precision=1},
enlarge x limits=.16]
\addplot[fill=echoblue!35,draw=none] coordinates {(1,98.044776) (2,96.019900) (3,96.019900) (4,94.029851) (5,71.641791)};
\addplot[fill=echoblue,draw=none,nodes near coords={}] coordinates {(1,98.044776)};
\end{axis}\end{tikzpicture}
\begin{tikzpicture}
\begin{axis}[width=.325\textwidth,height=4.65cm,ybar,bar shift=0pt,bar width=14pt,ymin=0,ymax=140,
xtick={1,2,3,4,5},xticklabels={ECHO,Mem0,A-MEM,Letta,LangMem},
xticklabel style={font=\tiny,rotate=25,anchor=east},yticklabel style={font=\scriptsize},
ylabel={Milliseconds},ylabel style={font=\scriptsize},title={(c) Mean retrieval latency},title style={font=\small},
ymajorgrids=true,grid style={black!10},axis line style={black!30},
nodes near coords,point meta=y,nodes near coords style={font=\tiny,/pgf/number format/fixed,/pgf/number format/precision=1},
enlarge x limits=.16]
\addplot[fill=echoblue!35,draw=none] coordinates {(1,19.541938) (2,120.263024) (3,20.513111) (4,84.172180) (5,29.455684)};
\addplot[fill=echoblue,draw=none,nodes near coords={}] coordinates {(1,19.541938)};
\end{axis}\end{tikzpicture}

\caption{Latest supplied aggregate comparison. All axes begin at zero. Panels (a) and (b) favor higher values; panel (c) favors lower values and measures retrieval only. No uncertainty bars are shown because question-level records for this snapshot were not supplied.}
\label{fig:comparison}
\end{figure*}

\subsection{Retrieval Latency}
\begin{table}[t]
\caption{Reported retrieval latency in milliseconds. ECHO's P50/P95 entries are transposed from the supplied spreadsheet according to the authors' correction. Generation latency is not supplied.}
\label{tab:latency}
\centering\small
\begin{tabular}{lrrr}
\toprule
System & Mean & P50 & P95 \\
\midrule
ECHO & 19.54 & 19.12 & 19.43 \\
Mem0 & 120.26 & 117.71 & 215.12 \\
A-MEM & 20.51 & 20.35 & 21.80 \\
Letta & 84.17 & 83.70 & 90.17 \\
LangMem & 29.46 & 23.96 & 56.35 \\
\bottomrule
\end{tabular}
\end{table}
ECHO's reported mean retrieval latency is 19.54\,ms, compared with 120.26\,ms for Mem0, 20.51\,ms for A-MEM, 84.17\,ms for Letta, and 29.46\,ms for LangMem. The Mem0-to-ECHO mean ratio is approximately 6.15 in this snapshot. The table does not establish matching hardware, concurrency, cache state, embedding inclusion, or request boundaries, so this ratio is not an end-to-end speedup claim.

The corrected ECHO P50 and P95 values are 19.12 and 19.43\,ms. A mean above P95 is mathematically possible because of a heavy upper tail; the raw timing distribution is required to assess that explanation. We retain the supplied mean and the explicitly corrected percentiles without inventing a tail distribution. No generation or ingestion timing is inferred from the retrieval values.

\subsection{Relationship to Earlier Published Evidence}
The earlier public version \cite{qian2026echo} reported a different experimental lineage. Its expansion-enabled development retrieval reached 96.29\% Hit@10 on 1,536 LoCoMo category-1--4 questions and 97.60\% on 500 LongMemEval-S questions. A post-hoc audit found source-specific phrases in expansion rules, limiting those results to development evidence. Its five-history BEAM gate failed, with 78.65\% Hit@10, 40.80\% turn recall, and 52.34\% session recall on 89 evaluated questions.

The same public version reported a separate matched 91-question QA sample: ECHO answered 38 correctly (41.76\%) and Mem0 59 (64.84\%). The question-level exact McNemar result was $p=.00107$, while a history-cluster sensitivity interval crossed zero. These earlier results remain historical evidence; the latest spreadsheet neither reruns that sample nor removes its limitations. Changes in systems, readers, data, and weighting prevent interpreting the difference between old and new aggregate percentages as a measured longitudinal improvement.

\section{Related Work}
Persistent memory systems differ in what they retain and how they expose it to an agent. Generative Agents uses observations and reflection \cite{park2023generative}; MemGPT manages context across memory tiers \cite{packer2023memgpt}; A-MEM links evolving notes \cite{xu2025amem}; and Mem0 extracts and updates salient memory \cite{chhikara2025mem0}. Temporal graph architectures such as Zep make time and relations explicit \cite{rasmussen2025zep}. These establish substantial prior art for memory organization. ECHO's emphasis is an auditable separation between discovery, revision authority, source packing, and answering.

Temporal reasoning also requires more than storing timestamps. TReMu combines timeline memory with explicit calculation \cite{ge2025tremu}; LightMem studies filtering and consolidation \cite{fang2025lightmem}; and graph-based approaches support associative retrieval \cite{gutierrez2025hipporag2}. ECHO builds on temporal database and provenance concepts \cite{torp2000timestamping,cheney2009provenance} rather than claiming that bitemporal storage or source tracing is novel in isolation.

Evaluation must distinguish stages and tasks. MemoryAgentBench includes learning, long-range understanding, and forgetting \cite{hu2025memoryagentbench}; HaluMem distinguishes memory hallucinations across lifecycle stages \cite{chen2025halumem}. Lost in the Middle shows that supplying context does not ensure its effective use \cite{liu2024lost}. Together these motivate separate coverage, support, correctness, and latency measures. Scores reported under different readers, judges, contexts, and provenance mappings are not directly interchangeable.

\section{Discussion and Limitations}
\subsection{What the Current Snapshot Supports}
The latest table is a promising development snapshot, with favorable quality and retrieval cost across its five systems. It does not establish full-dataset SOTA, production readiness, or the independent contribution of an architectural component.

A particularly important distinction is between traceability and truth. A system can faithfully trace an incorrect projection to a source, or cite a relevant record while making an unsupported inference. Immutable evidence and revisions make such errors inspectable; they do not eliminate them. Likewise, turning off model reranking is a concrete configuration choice, not an experimentally isolated explanation for the latest accuracy or latency.

\subsection{Reproducibility Gaps and Next Experiments}
The next evaluation should freeze the manifests and per-question artifacts identified in Section~\ref{sec:eval}, rerun all five systems under a shared reader and judge, and publish failures alongside successful outputs. These records are necessary for independent reproduction and paired uncertainty estimates.

Ablations should separately vary retrieval routes, reranking, original-source packing, evidence-use checks, and revision resolution, holding all other inputs constant. The present snapshot is not such an ablation. Temporal tests should include late evidence, corrections, revocations, ambiguous entities, and historical queries. Latency tests should separate cold and warm retrieval, query embedding, packing, reader generation, and complete wall-clock requests on named hardware.

The historical BEAM failure and expansion audit motivate source-neutral query transformations and a fresh confirmatory split. The latest aggregate cannot demonstrate that those issues have been eliminated. Human calibration of correctness and hallucination judges is also needed, especially for preferences, summaries, and abstention. Larger contexts and higher source recall are diagnostic improvements only when accompanied by better supported answers.

\subsection{Embodied Use, Privacy, and Governance}
The architecture accepts perception-derived events and provides action and feedback hooks, but the reported metrics concern memory retrieval and answer quality. They do not demonstrate multimodal grounding, proactive assistance, autonomous learning, or safe physical action. These require separate situated evaluations with controlled interventions and action-level outcomes.

Persistent personal memory creates access-control, consent, retention, and deletion obligations. Subject and robot scope must constrain queries and revisions, while deletion or revocation must propagate through derived indexes and cached packets. Auditability should not become indefinite retention of sensitive source material. Shared artifacts should exclude credentials, private conversations, infrastructure addresses, and unnecessary personal identifiers. The mechanism of evidence traceability is compatible with these goals only when governance covers both the source store and its derived views.

\section{Conclusion}
ECHO organizes long-term memory as a dual-path lifecycle: the fast path preserves events and serves recall, while the slow path forms and revises source-linked state. Explicit boundaries separate discovery, temporal authority, evidence packing, and answer realization. The latest author-supplied aggregate snapshot reports 56.27\% answer accuracy using mixed original-dialogue and memory-summary inputs with a Qwen3-8B NF4 reader, and 19.54\,ms mean retrieval latency, with favorable within-table comparisons across five systems. These observations remain distinct from engineering correctness and earlier published experiments. Matched per-question artifacts, fresh-history validation, and separately measured generation latency are required to turn this development snapshot into a reproducible comparative result.

\appendices
\section{Revision and Data Provenance}
The authors designated the latest five-system workbook as authoritative and corrected ECHO's P50/P95 cells by exchanging them. Other values are unchanged. Derived differences and ratios use unrounded inputs; tables display rounded values. The source package includes the corrected machine-readable aggregates and a separate reader-configuration disclosure. The latter records the authors' subsequent confirmation of mixed original-dialogue/summary inputs and Qwen3-8B NF4, without claiming that missing runtime fields have been independently verified.

The original workbook and pre-edit manuscript are retained in the revision archive. Earlier experiments remain separate. No success counts, confidence intervals, or per-dataset scores have been reconstructed from aggregate percentages.

\bibliographystyle{IEEEtran}
\bibliography{references}
\end{document}